\def\year{2021}\relax
\documentclass[letterpaper]{article} 
\usepackage{aaai21}  
\usepackage{times}  
\usepackage{helvet} 
\usepackage{courier}  
\usepackage[hyphens]{url}  
\usepackage{graphicx} 
\usepackage{amsmath}
\usepackage{amssymb}
\usepackage{multirow}
\usepackage{amsmath,bm,amsfonts}
\usepackage{diagbox}
\usepackage{booktabs}
\usepackage{color}
\usepackage{array}
\usepackage[switch]{lineno}
\newcommand{\PreserveBackslash}[1]{\let\temp=\\#1\let\\=\temp}
\newcolumntype{C}[1]{>{\PreserveBackslash\centering}p{#1}}
\newcolumntype{R}[1]{>{\PreserveBackslash\raggedleft}p{#1}}
\newcolumntype{L}[1]{>{\PreserveBackslash\raggedright}p{#1}}
\usepackage{natbib}  
\usepackage{caption} 
\title{Image Captioning with Context-Aware Auxiliary Guidance}

\author {
	Zeliang Song,\textsuperscript{\rm 1,2}
	Xiaofei Zhou,\textsuperscript{\rm 1,2}\thanks{Corresponding Author}
	Zhendong Mao,\textsuperscript{\rm 3} 
	Jianlong Tan\textsuperscript{\rm 1,2} 
	\\
}
\affiliations {
	\textsuperscript{\rm 1}Institute of Information Engineering, Chinese Academy of Sciences, Beijing, China \\
	\textsuperscript{\rm 2}School of Cyber Security, University of Chinese Academy of Sciences, Beijing, China \\
	\textsuperscript{\rm 3}University of Science and Technology of China, Hefei, China \\
	\{songzeliang, zhouxiaofei\}@iie.ac.cn, zdmao@ustc.edu.cn
}

\begin{document}
	
	\maketitle
	
	\begin{abstract}
		Image captioning is a challenging computer vision task, which aims to generate a natural language description of an image. Most recent researches follow the encoder-decoder framework which depends heavily on the previous generated words for the current prediction. Such methods can not effectively take advantage of the future predicted information to learn complete semantics. In this paper, we propose  Context-Aware Auxiliary Guidance (CAAG) mechanism that can guide the captioning model to perceive global contexts. Upon the captioning model, CAAG performs semantic attention that selectively concentrates on useful information of the global predictions to reproduce the current generation. To validate the adaptability of the method, we apply CAAG to three popular captioners and our proposal achieves competitive performance on the challenging Microsoft COCO image captioning benchmark, e.g. 132.2 CIDEr-D score on Karpathy split and 130.7 CIDEr-D (c40) score on official online evaluation server. 
	\end{abstract}

	\section{Introduction}
	Image captioning is to automatically generate natural language descriptions of an input image, which is a challenging task and also draws increasing attention in both computer vision and natural language processing fields. The ability of describing like humans for machine is important since it can be widely applied to cross-modal retrieval \cite{ITS,CTI,LTB} and human-robot interactions \cite{NVD,HID,ASD}.
	
	Most recent image captioning approaches follow the encoder-decoder framework \cite{SAT} which employs convolutional neural network (CNN) to encode images to visual features and utilizes  recurrent neural network (RNN) decoder to generate captions. Inspired by the recent success of employing attention in machine translation \cite{NMT}, the spatial attention mechanism is proposed in \cite{SAAT} to attend to the salient regions of an image while generating captions. These methods are typically trained by ``Teacher-Forcing'' algorithm \cite{TF} to maximize the log-likelihood of the next ground-truth word. This training approach creates a mismatch between training and inference called ``exposure bias'' \cite{Exposure}. During the training phase, the model takes the ground-truth word as the current input, while in the inference phase, the input can be only sampled from the last time step. Recently, it has shown that REINFORCE algorithm \cite{reinforce} can avoid exposure bias through directly optimizing non-differentiable sequence-level metrics, and has become a common component in the field of image captioning.

	\begin{figure}[t]
		\centering
		\includegraphics[width=0.95\columnwidth]{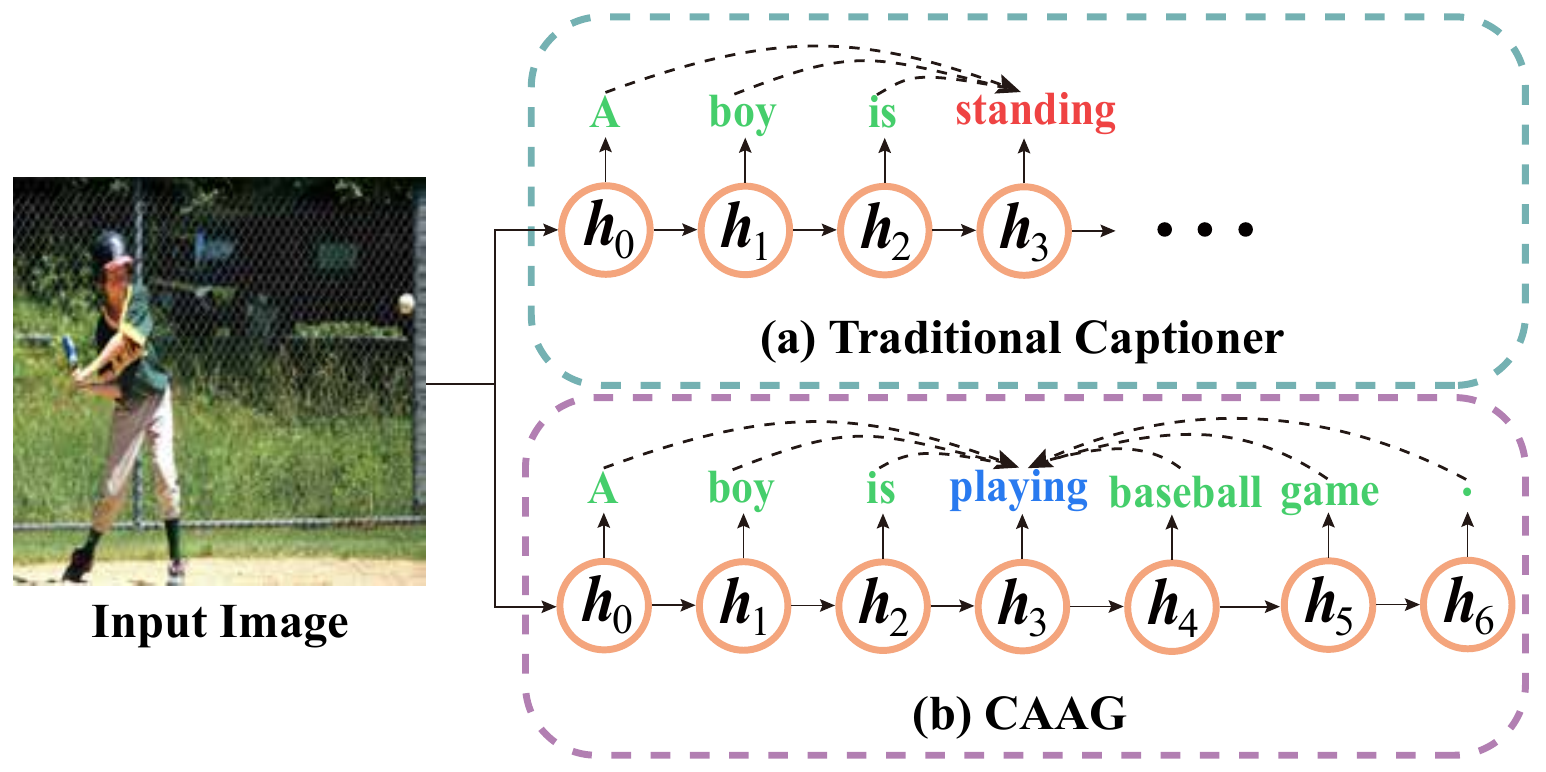}
		\caption{Comparison between (a) traditional captioner and (b) our CAAG. The traditional captioner depends solely on the previous generated words when predicting the current word. Our method can take advantage of the future predictions to guide model learning.}
		\label{motivation}
	\end{figure}
	
	However, the existing image captioning methods depend solely on the previous generated words when conducting the current prediction, which will lead to the model learning incomplete semantics. As illustrated in Figure \ref{motivation}, the traditional captioner predicts the current word based on the partial generated sentence ``A boy is'' during training. With the incomplete information, the caption decoder tends to learn by remembering experiences rather than by understanding the scenes, and will easily generate incorrect word (e.g. ``standing'') when encountering similar situations in the inference stage. 
	Therefore, the future predictions ``baseball game'' may contain more critical information for current word, and should be considered to improve the scene understanding ability of captioning model. 
	
	In order to better understand the scenes, we propose  Context-Aware  Auxiliary Guidance (CAAG) mechanism that guides the captioning model to perceive the complete semantics through reproducing the current generation based on the global predictions. We first use the captioning model (called primary network) to generate a complete sentence which is served as global contexts. Based on the global contexts and hidden state, CAAG masks and then reproduces the target word through semantic attention at every time step. The semantic attention can help CAAG to selectively look backward or look into future. During the inference phase, we first utilize primary network to generate a sentence for CAAG by greedy decoding. Then we jointly employ primary network and CAAG to generate the final caption by beam search.
	
	We conduct extensive experiments and analyses on the challenging Microsoft COCO image captioning benchmark to evaluate our proposed method. To validate the adaptability of our method, we apply CAAG to three popular image captioners (Att2all \shortcite{SCST}, Up-Down \shortcite{BTATD} and AoANet \shortcite{AOA}), and our  model achieves consistent improvements over all metrics. 
	
	The major contributions of our paper can be summarized as follows: 
	\begin{itemize}
		\item We propose  Context-Aware Auxiliary Guidance (CAAG) mechanism that guides the captioning model to perceive more complete semantics through taking advantage of the future predicted information to enhance the ability of model learning for image captioning.
		\item Our proposed method is generic so that it can enhance existing reinforcement learning based image captioning models and we show consistent improvements over three typical attention based LSTM decoders
		by experiments.
		\item Our model outperforms many state-of-the-art methods on  the challenging Microsoft COCO dataset. More remarkably, CAAG improves CIDEr-D performance of a top-down attention based LSTM decoder from 123.4\% to 128.8\% on Karpathy test split.
	\end{itemize} 
	
	\begin{figure*}[t]
		\centering
		\includegraphics[width=1.8\columnwidth]{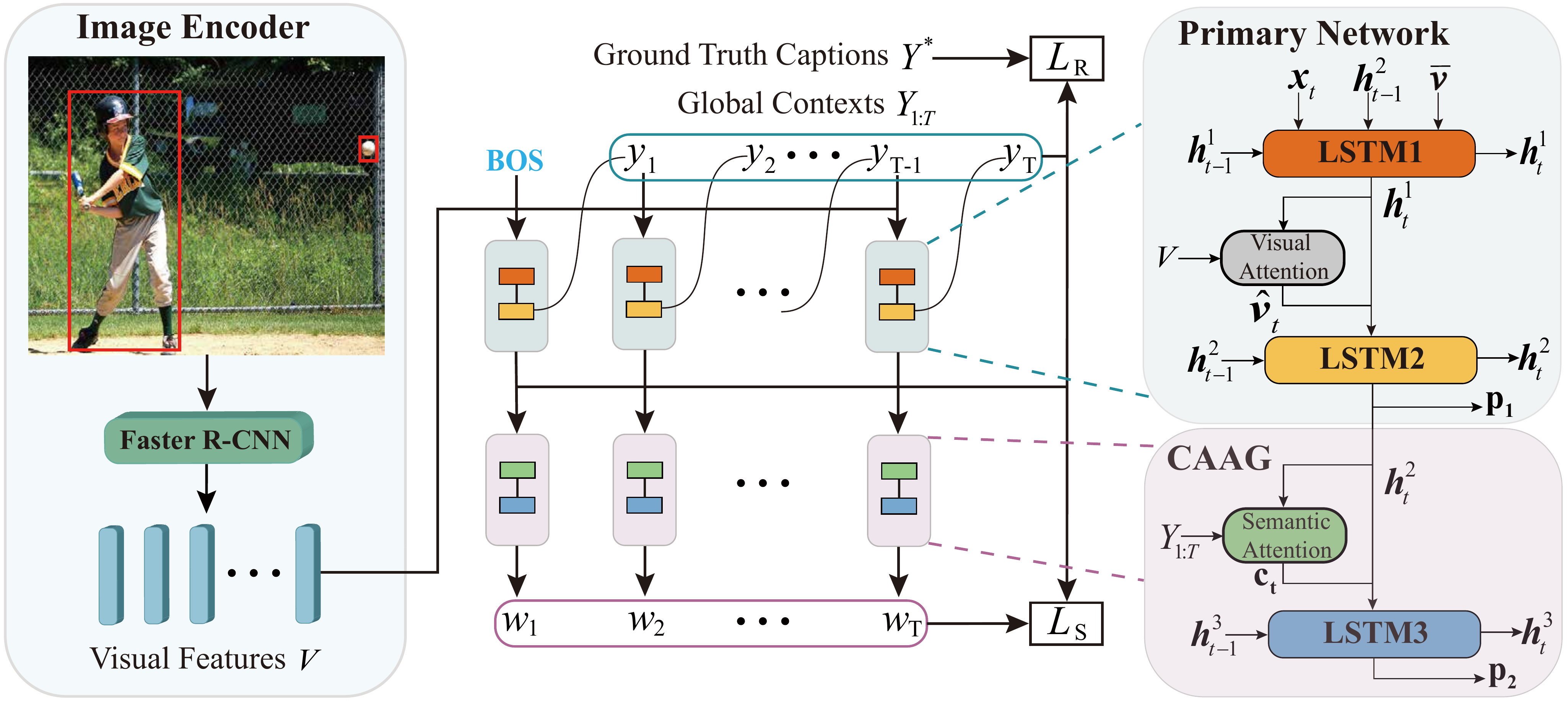}
		\caption{The overall workflow of our proposed method. Firstly, we employ Faster R-CNN to extract spatial visual features of salient regions. Secondly, the primary network generates a complete sentence (global contexts) based on the visual features. Thirdly, we utilize CAAG to guide the primary network to perceive the global contexts.}
		\label{workflow}
	\end{figure*}
	
	\section{Related Work}
	In recent years, the literatures on image captioning have grown rapidly, which can be divided into two categories: bottom-up methods \cite{IDVD,FROM,CGN,CSD,MID} and top-down methods \cite{SAT,ICSA,NBT,DA,RNCG}. Top-down methods achieve state-of-the-art performance, and they follow the encoder-decoder framework \cite{SAT} which employs a pre-trained CNN to encode the input image into feature vectors and utilizes a RNN to decode these vectors into captions. In this section, we mainly introduce recent methods of this branch. Attention mechanism \cite{SCA,knowing} and reinforcement learning algorithm have been proved to have significant improvements on caption generation and have been widely applied to top-down methods.
	
	\subsection{Attention Mechanism}
	Inspired by recent work in machine translation and object detection, Kelvin Xu et al. \shortcite{SAAT} first introduce soft and hard attention mechanisms to focus on different salient regions of an image at different time steps. Soft attention takes the weighted average of value vectors as attention result, while hard attention samples a value vector according to the relevance weights. Anderson et al. \shortcite{BTATD} take pre-trained Faster R-CNN on Visual Genome dataset \cite{VG} as a feature extractor to improve attentive models by replacing the attention over a grid of features with attention over image regions. Guo et al. \shortcite{STP} design ruminant decoder to polish the raw caption generated by the base decoder to obtain a more
	comprehensive caption. Yao et al. \shortcite{explore} novelly integrate both semantic and spatial object relationships into image encoder. Yu et al. \shortcite{LBPF} present Predict Forward (PF) model that predicts the next two words in one time step to embed more precise information into hidden states.  Herdade  et al. \shortcite{transform} introduce the object relation transformer for image captioning. Li et al. \shortcite{ETAR} propose EnTangled Attention (ETA) that enables the transformer to exploit semantic and visual information simultaneously. Yao et al. \shortcite{HIP} integrate hierarchical structure into image encoder. Huang et al. \shortcite{AOA} extend the traditional attention mechanism by determining the relevance of attention results.
	
	\subsection{Reinforcement Learning}
	Recently, it has shown that reinforcement learning algorithm can avoid the exposure bias \cite{Exposure} problem and can directly optimize non-differentiable evaluation metrics, such as BLEU \cite{BLEU}, ROUGE \cite{ROUGE}, METEOR \cite{METEOR}, CIDEr \cite{CIDEr} and SPICE \cite{SPICE}.  Siqi Liu et al. \shortcite{SPIDER} directly optimize SPIDEr (a linear combination of SPICE and CIDEr) and use different Monte Carlo rollouts to get a better estimate of the value function. In \cite{TD}, they propose temporal difference method for image captioning where each action at different time step has different impacts on the model. Junlong Gao et al. \shortcite{SCSTN} extend to \emph{n}-step reformulated advantage function and use different kinds of rollouts to estimate the state-action value function. In \cite{SCST}, they present self-critical sequence training (SCST) algorithm that utilizes the output of greedy decoding to normalize the reward it experiences rather than estimating the reward signal. SCST has become the most popular training method, because it only needs one additional forward propagation.

	\section{Architecture}
	Like most existing methods \cite{BTATD,CAVP,LBPF}, we utilize a pre-trained Faster R-CNN on Visual Genome dataset to extract a variably-sized set of $k$ spatial features $\bm{V}=\{\bm{v}_1,\bm{v}_2,\cdots,\bm{v}_k\}$ for an input image $I$, where $\bm{v}_i\in\mathbb{R}^{d_I}$, and $d_I=2048$.  The primary network firstly generates a sentence $Y_{1:T}=\{y_1,y_2,\cdots,y_T\}$ as the global contexts.  CAAG performs semantic attention based on the global contexts to guide the primary network learning. Figure \ref{workflow} shows the architecture of CAAG and primary network.  To explain the working mechanism of CAAG clearly, we select the typical Up-Down model \cite{BTATD} as the primary network.
	
	\subsection{Primary Network}
	We employ the classic Up-Down captioner \cite{BTATD} as our primary network, which utilizes adaptive attention to dynamically attend to spatial features through an attention LSTM (LSTM1) and a language LSTM (LSTM2). At time step $t$, the input vector to LSTM1 consists of the embedding $\bm{x}_t$ of previous generated word, the previous hidden state $\bm{h}_{t-1}^2$ of LSTM2 and the mean-pooled image feature $\bar{\bm{v}}=\frac{1}{k}\sum_{i}\bm{v}_i$. Given the output $\bm{h}_{t}^1$ of LSTM1, we calculate a weighted average vector $\hat{\bm{v}}_t$ as follows:
	\begin{equation}\label{visual-attention}
	\hat{\bm{v}}_t=\sum_{i=1}^{k}\alpha_{i,t}\bm{v}_i
	\end{equation}
	where the normalized weight $\alpha_{i,t}=f_{att}(\bm{v}_i, \bm{h}_t^1)$ for each spatial feature $\bm{v}_i$ is given by:
	\begin{gather}\label{alpha-weight}
	u_{i,t}=\bm{w}_u^T\tanh(\bm{W}_{vu}\bm{v}_i+\bm{W}_{hu}\bm{h}^1_t) \\
	\bm{\alpha}_t=softmax(\bm{u}_t)
	\end{gather}
	and $\bm{W}_{vu}$, $\bm{W}_{hu}$, and $\bm{w}_u$ are learned parameters of $f_{att}$, $\bm{\alpha}_t=\{\alpha_{1,t},\alpha_{2,t},\cdots,\alpha_{k,t}\}$. 
	
	The input vector to LSTM2 consists of the attended spatial feature $\hat{\bm{v}}_t$, concatenated with the output $\bm{h}_t^1$ of LSTM1. Based on the output $\bm{h}_t^2$ of LSTM2, the conditional distribution over the dictionary is given by:
	\begin{equation}\label{primary-policy}
	p^1_t(y_{t+1}|Y_{1:t})=softmax(Linear(\bm{h}_t^2))
	\end{equation}
	where $p^1_t$ denotes the output probability distribution of primary network and $Y_{1:t}$ denotes the partially generated sentence.
	
	\subsection{Context-Aware Auxiliary Guidance}
	Given the global contexts $Y_{1:T}$ generated by primary network. At time step $t$, based on the word embedding $Y_{emb}=\{\bm{x}_1,\bm{x}_2,\cdots,\bm{x}_T\}$ of  $Y_{1:T}$ and the hidden state $\bm{h}^2_t$,  CAAG performs following semantic attention to generate a contextual vector $\bm{c}_{t}$. 
	\begin{gather}\label{location-weight}
	\beta_{i,t}=f_{att}^2(\bm{x}_{i}, \bm{h}_t^2) \\
	\bm{c}_{t}=\sum_{i=1}^{T}\beta_{i,t}\bm{x}_i
	\end{gather}
	where $f^2_{att}$ is the attention function and $\beta_{i,t}$ is the normalized attention weight.  Note that the target word $y_{t+1}$ is masked inside the semantic attention mechanism to be “self-unknown” during the training stage. 
	
	The input vector to LSTM3 consists of the contextual vector $\bm{c}_{t}$, concatenated with the output $\bm{h}_t^2$ of LSTM2. Based on the output $\bm{h}_t^3$ of LSTM3, the conditional distribution is given by:
	\begin{equation}\label{auxiliary-policy}
	p^2_t(y_{t+1}|Y_{1:T})=softmax(Linear(\bm{h}_t^3))
	\end{equation}
	where $p^2_t$ denotes the output probability distribution of auxiliary network.
	In the inference phase, we jointly apply the output probability distributions $p^1_t$  and $p^2_t$ to generate the captions. The final probability distribution is given by:
	\begin{equation}\label{final-policy}
	p_t=p_t^1+\lambda p_{t}^2
	\end{equation}
	where $\lambda$ is a trade-off coefficient.
	
	\section{Objectives}
	In this section, we will introduce the objective functions of primary network and CAAG respectively.
	
	\subsection{Objective of Primary Network}
	Given the ground truth sentence $Y^*_{1:T}$ and the parameters $\bm{\theta}_p$ of primary network, we pre-train the model by minimizing the following cross entropy loss:
	\begin{equation}
	L_{C}(\bm{\theta}_p)=-\sum_{t=0}^{T-1}\log (p^1_t(y_{t+1}^*|Y_{1:t}^*))
	\end{equation}
	
	To address the exposure bias problem of the cross entropy loss, we consider image captioning as a sequential decision making problem. Specifically, the primary network can be viewed as an ``agent" which interacts with external ``environment" (input tokens, spatial features). The ``action" is the prediction of the next word through ``policy" $p^1_t$. The ``state" can be simply viewed as the current hidden state $\bm{h}_t^2$ of LSTM2. After generating the whole sentence $Y_{1:T}=\{y_1,y_2,...,y_T\}$, the  ``agent" observes a ``reward" $r$, i.e. a language evaluation metric score (CIDEr, BLEU, SPICE, etc.) computed between $Y_{1:T}$ and corresponding ground truth captions. The ``goal" of the agent is to maximize the following expected reward:
	\begin{equation}\label{reward}
	L_R(\bm{\theta}_p)=\mathbb{E}_{Y_{1:T}\sim p_1}[r(Y_{1:T})]
	\end{equation}
	Following the REINFORCE algorithm \cite{reinforce}, the expected gradient can be approximated using a single Monte-Carlo sample $Y_{1:T}\sim p_1$:
	\begin{equation}\label{gradient}
	\frac{\partial L_R(\bm{\theta}_p)}{\partial \bm{\theta}_p} \approx \sum_{t=0}^{T-1}A_t\frac{\partial \log p^1_t(y_{t+1}|Y_{1:t})}{\partial \bm{\theta}_p}
	\end{equation} 
	where $A_t=r(Y_{1:t})-b$ is called advantage function and $b$ denotes the baseline function that can reduce the variance of the gradient estimate without changing its expectation.
	
	\subsection{Objective of CAAG}
	Given the sentence $Y_{1:T}$ generated by primary network and the parameters $\bm{\theta}_a$ of CAAG, we maximize the following objective function:
	\begin{equation}\label{caag}
	L_{S}(\bm{\theta}_a)=\sum_{t=0}^{T-1}\log (p^2_t(y_{t+1}|Y_{1:T}))A_t
	\end{equation}
	where $A_t$ increases the probability of the samples with higher reward than greedy decoding and suppresses samples with lower reward.
	
	During the training phase, the parameters of the primary network and CAAG are trained simultaneously, i.e. we maximize the following objective function:
	\begin{equation}
	L(\bm{\theta}_p,\bm{\theta}_a) = L_R(\bm{\theta}_p)+\gamma L_S(\bm{\theta}_a)
	\end{equation}
	
	\section{Experiments}
	In this section, we first describe the datasets and settings of our experiments. Then, we go through the quantitative analysis and ablation studies. Finally, we introduce the qualitative analysis and human evaluation.
	\subsection{Datasets and Settings}
	\subsubsection{Visual Genome Dataset}
	Visual Genome dataset \cite{VG} contains 108,077 images, 3.8 million object instances, 2.8 million attributes, and 2.3 million pairwise relationships between objects. Every image includes an average of 42 regions with a bounding box and a descriptive phrase. In this paper, we employ Faster R-CNN pre-trained \cite{BTATD} on Visual Genome dataset, which is split with 98K / 5K / 5K images for training/validation/testing, to extract spatial features. Notice that only the object and attribute data are used during pre-training.
	
	\subsubsection{Microsoft COCO Dataset}
	We evaluate our proposed method on the Microsoft COCO (MSCOCO) 2014 captions dataset \cite{MSCOCO}. MSCOCO contains totally 164,062 images and is split with 2:1:1 for training, validation and testing. Each image in MSCOCO is given at least 5 ground truth captions by different AMT workers. For hyperparameters selection and offline evaluation, we use the publicly available Karpathy split\footnote{\url{https://github.com/karpathy/neuraltalk}} which contains 113,287 training images, and 5K images respectively for  validation and testing. For online evaluation on the MSCOCO test server, we add the 5K testing set into the training set to form a larger training set (118,287 images). We truncate all the sentences in training set to ensure that any sentence does not exceed 16 characters.  We follow standard practice and perform only minimal text pre-processing \cite{BTATD}: tokenizing on white space, converting all sentences to lower case, and filtering words that occurs less than five times, resulting in a vocabulary of 10,096 words. 
	
	\subsubsection{Evaluation Metrics} To evaluate the quality of the generated captions, we use MSCOCO caption evaluation tool\footnote{\url{https://github.com/tylin/coco-caption}} to calculate standard evaluation metrics, including BLEU \cite{BLEU}, METEOR \cite{METEOR}, ROUGE \cite{ROUGE}, CIDEr \cite{CIDEr} and SPICE \cite{SPICE}.

	\begin{table*}[t]
		\centering
		\scalebox{0.93}{\begin{tabular}{lcccccccccccccc}
				\hline
				\multirow{2}*{Models} & \multicolumn{2}{c}{BLEU-1} & \multicolumn{2}{c}{BLEU-2} &
				\multicolumn{2}{c}{BLEU-3} & 
				\multicolumn{2}{c}{BLEU-4} &
				\multicolumn{2}{c}{METEOR} &
				\multicolumn{2}{c}{ROUGE-L} & 
				\multicolumn{2}{c}{CIDEr-D} \\
				\cmidrule(lr){2-3} \cmidrule(lr){4-5} \cmidrule(lr){6-7} \cmidrule(lr){8-9} \cmidrule(lr){10-11} \cmidrule(lr){12-13} \cmidrule(lr){14-15}
				& c5 & c40 & c5 & c40 & c5 & c40 & c5 & c40 & c5 & c40 & c5 & c40 & c5 & c40
				\\
				\hline
				Google NIC \shortcite{SAT} & 71.3 & 89.5 & 54.2 & 80.2 & 40.7 & 69.4 & 30.9 & 58.7 & 25.4 & 34.6 & 53.0 & 68.2 & 94.3 & 94.6 \\
				
				SCST \shortcite{SCST} & 78.1 & 93.7 & 61.9 & 86.0 & 47.0 & 75.9 & 35.2 & 64.5 & 27.0 & 35.5 & 56.3 & 70.7 & 114.7 & 116.7 \\
				
				Up-Down \shortcite{BTATD} & 80.2 & 95.2 & 64.2 & 88.8 & 49.1 & 79.4 & 36.9 & 68.5 & 27.6 & 36.7 & 57.1 & 72.4 & 117.9 & 120.5 \\
				
				CAVP \shortcite{CAVP} & 80.1 & 94.9 & 64.7 & 88.8 & 50.0 & 79.7 & 37.9 & 69.0 & 28.1 & 37.0 & 58.2 & 73.1 & 121.6 & 123.8 \\
				GCN-LSTM \shortcite{explore} & 80.8 & 95.9 & 65.5 & 89.3 & 50.8 & 80.3 & 38.7 & 69.7 & 28.5 & 37.6 & 58.5 & 73.4 & 125.3 & 126.5 \\
				
				ETA \shortcite{ETAR} & 81.2 & 95.0 & 65.5 & 89.0 & 50.9 & 80.4 & 38.9 & 70.2 & 28.6 & 38.0 & 58.6 & 73.9 & 122.1 & 124.4 \\
				SGAE \shortcite{AE} & 81.0 & 95.3 & 65.6 & 89.5 & 50.7 & 80.4 & 38.5 & 69.7 & 28.2 & 37.2 & 58.6 & 73.6 & 123.8 & 126.5 \\
				
				AoANet \shortcite{AOA} & 81.0 & 95.0 & 65.8 & 89.6 & 51.4 & 81.3 & 39.4 & 71.2 & 29.1 & 38.5 & 58.9 & 74.5 & 126.9 & 129.6 \\
				GCN+HIP \shortcite{HIP} & \textbf{81.6} & \textbf{95.9} & 66.2 & 90.4 & 51.5 & 81.6 & 39.3 & 71.0 & 28.8 & 38.1 & 59.0 & 74.1 & 127.9 & 130.2 \\
				
				\hline
				
				\textbf{Ours} & 81.1 & 95.7 & \textbf{66.4} & \textbf{90.5} &\textbf{51.7} & \textbf{82.3} & \textbf{39.6} &  \textbf{72.0} & \textbf{29.2} & \textbf{38.6} & \textbf{59.2} & \textbf{74.7} & \textbf{128.3} & \textbf{130.7} \\
				\hline
		\end{tabular}}
		\caption{Leaderboard of the published state-of-the-art image captioning models on the official online MSCOCO evaluation server. Notice that CIDEr-D is the most important metric which shows high agreement with consensus as assessed by humans. All results are reported in percentage (\%), with the highest score of each entry marked in boldface.}
		\label{online}
	\end{table*}
	\begin{table}
		\begin{center}
			\scalebox{0.9}{\begin{tabular}{l|c|c|c|c|c}
					\hline
					Models   & B-4 & M & R & C & S\\
					\hline
					SCST(Att2all) \shortcite{SCST} & 34.2 & 26.7 & 55.7 & 114.0 & -- \\
					
					Up-Down \shortcite{BTATD}   & 36.3 & 27.7 & 56.9 & 120.1 & 21.4 \\
					
					CAVP \shortcite{CAVP}   & 38.6 & 28.3 & 58.5 & 126.3 & 21.6\\
					GCN-LSTM \shortcite{explore}  & 38.3 & 28.6 & 58.5 & 128.7 & 22.1\\
					
					LBPF \shortcite{LBPF}   & 38.3 & 28.5 & 58.4 & 127.6 & 22.0\\
					
					SGAE \shortcite{AE}  & 38.4 & 28.4 & 58.6 & 127.8 & 22.1\\
					
					GCN+HIP \shortcite{HIP}  & 39.1 & 28.9 & 59.2 & 130.6 & 22.3\\
					ETA \shortcite{ETAR}  & 39.3 & 28.8 & 58.9 & 126.6 & 22.7\\
					AoANet \shortcite{AOA}  & 39.1 & 29.2 & 58.8 & 129.8 & 22.4\\
					\hline
					Att2all+CAAG         & 36.7& 27.9 & 57.1 & 121.7 & 21.4       \\
					Up-Down+CAAG & 38.4 & 28.6 & 58.6 & 128.8 & 22.1\\
					
					AoANet+CAAG  & \textbf{39.4} & \textbf{29.5} & \textbf{59.2} & \textbf{132.2} & \textbf{22.8} \\
					\hline
			\end{tabular}}
		\end{center}
		\caption{Performance comparisons of various methods on MSCOCO Karpathy split. The best results for each metric are marked in boldface separately. The metrics: B-4, M, R, C, and S denote BLEU-4, METEOR, ROUGE-L, CIDEr-D, SPICE respectively.}
		\label{offline}
	\end{table}
	
	\subsubsection{Implementation Details}
	We employ Up-Down captioner as our primary network, so we use the same hyperparameters proposed in \cite{BTATD} for fair comparison. The Faster R-CNN implementation uses an IoU threshold of 0.7 for region proposal suppression, 0.3 for object class suppression, and a class detection confidence threshold of 0.2 for selecting salient image regions. For captioning model, we set the dimension of hidden states in both LSTMs to 1000, the number of hidden units in all attention layers to 512, and the dimension of input word embedding to 1000.  The trade-off coefficient in Eq. \ref{final-policy} is set to 0.5 and the batch size is 64. We use beam search with a beam size of 3 to generate captions when validating and testing. We employ ADAM optimizer with an initial learning rate of $5e^{-4}$, and momentum of 0.9 and 0.999.  We evaluate the model on the validation set at every epoch and select the model with highest CIDEr score as the initialization for reinforcement learning. For self-critical learning, we select CIDEr score as our reward function. The learning rate starts from  $5e^{-5}$ and decays by rate 0.1 every 50 epochs.
	
	\subsubsection{Compared Methods} Although there are various image captioning approaches in recent years, for fair comparison, we only compare our method with some reinforcement learning based methods. To be specific, we compare our method with the following state-of-the-arts: \textbf{SCST(Att2all)} \cite{SCST}, \textbf{Up-Down} \cite{BTATD}, \textbf{CAVP} \cite{CAVP}, \textbf{GCN-LSTM} \cite{explore}, \textbf{LBPF} \cite{LBPF}, \textbf{SGAE} \cite{AE}, \textbf{GCN+HIP} \cite{HIP}, \textbf{ETA} \cite{ETAR} and \textbf{AoANet} \cite{AOA}. SCST and Up-Down are two baselines where the self-critical reward and the fine-grained spatial features are used. CAVP allows the previous visual features to serve as visual context for current decision. GCN-LSTM novelly integrates both semantic and spatial object relationships into image encoder. LBPF proposes Look-Back (LB) part to embed visual information from the past and Predict Forward (PF) part to look into future. SGAE uses the scene graph to represent the complex structural layout of both image and sentence. GCN+HIP integrates hierarchical structure into image encoder to enhance the instance-level, region-level and image-level features. ETA introduces EnTangled Attention to exploit semantic and visual information simultaneously and Gated Bilateral Controller to guide the interactions between the multimodal information. AoANet extends the conventional attention mechanisms to determine the relevance between attention results and queries. 
	
	\begin{table}[t]
		\centering
		\resizebox{.9\columnwidth}{!}{\begin{tabular}{l|c|c|c|c|c}
				\hline
				Models & B-4 & M & R & C & S\\
				\hline
				Base  & 36.9 & 28.0 & 57.5 & 123.4 & 21.5 \\
				\hline
				+RD \shortcite{STP} &37.8 & 28.2 & 57.9 & 125.3 & 21.7 \\
				+CAAG-P & 38.1 & 28.4 & 58.4 & 126.7 & 21.7 \\
				+CAAG-C & 38.2 & 28.5 & 58.4 & 127.2 & 21.8 \\
				+CAAG-H & 38.3 & 28.5 & 58.5 & 127.5 & 22.0 \\
				+CAAG & \textbf{38.4} & \textbf{28.6} & \textbf{58.6} & \textbf{128.8} & \textbf{22.1} \\
				\hline
		\end{tabular}}
		\caption{Settings and results of ablation studies on MSCOCO Karpathy split.}
		\label{ablation}
	\end{table}
	
	\subsection{Quantitative Analysis}

	\subsubsection{Offline Evaluation} The performance comparisons on Microsoft COCO Karpathy split are shown in Table \ref{offline}. To validate the generality and adaptability of our method, we implement CAAG over three popular image captioning models: Att2all \shortcite{SCST}, Up-Down \shortcite{BTATD} and AoANet \shortcite{AOA}. As is shown in Table \ref{offline}, the results indicate that our proposed method has a wide range of applicability to different image captioners. To be detailed, CAAG respectively makes the absolute improvement  over the baselines by 7.7\%, 8.7\% and 2.4\% in terms of CIDEr-D. Compared with the models (CAVP, GCN-LSTM, LBPF and SGAE) that directly extend Up-Down captioner, our Up-Down with CAAG achieves highest CIDEr-D score, which demonstrates the superiority of our method. Notice that LBPF proposes PF module to look ahead one step, and our method outperforms it over all metrics because CAAG can capture more powerful global contexts. When applying CAAG to AoANet, we outperform all the state-of-the-art methods (including transformer based ETA) across all evaluation metrics, and we obtain BLEU-4 / METEOR / ROUGE-L / CIDEr-D / SPICE scores of 39.4 / 29.5 / 59.2 / 132.2 / 22.8.
	
	\subsubsection{Online Evaluation} We also submit the run of AoANet+CAAG optimized with CIDEr-D reward to the official online testing server\footnote{\url{https://competitions.codalab.org/competitions/3221}}. The performance leaderboard on official testing dataset with 5 reference captions (c5) and 40 reference captions (c40) is shown in Table \ref{online}. Our results are evaluated by an ensemble of 4 models trained on the Karpathy  split for fair comparison. From the results shown in the table, GCN+HIP achieves better performance on BLEU-1 which is of little interest in recent years. Except for BLEU-1, our method outperforms all the state-of-the-art methods. For example, we achieve the highest score of 130.7 on the most important CIDEr-D (c40) metric which shows high agreement with consensus as assessed by humans.

	\begin{table}[t]
		\centering
		\scalebox{0.7}{\begin{tabular}{|L{0.15\linewidth}|L{0.5\linewidth}|L{0.5\linewidth}|}
				\hline
				&{\includegraphics[width=1\linewidth]{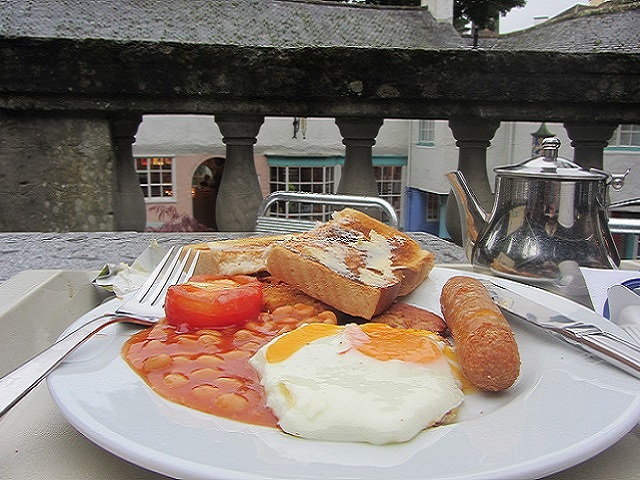}} &{\includegraphics[width=1\linewidth]{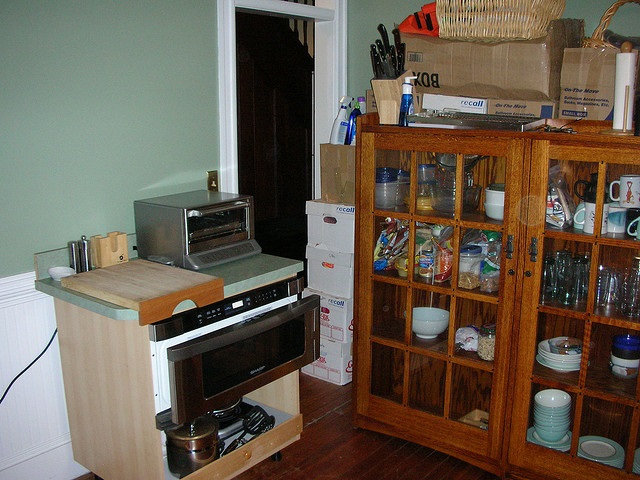}} \\
				\hline
				\multirow{2}*{Up-Down}&a plate of food on a table {\color{red} with a table}. & a kitchen with a stove and a {\color{red} refrigerator}. \\
				\hline
				\multirow{2}*{CAAG-P}& a plate of food on a table with a {\color{blue} fork} on it. &a kitchen with a stove and a {\color{blue} microwave}.\\
				\hline
				\multirow{2}*{CAAG}&a plate of food on a table with a {\color{blue} sandwitch} and a {\color{blue}fork}. & a kitchen with a stove and a {\color{blue} microwave} on a table. \\
				
				\midrule[1pt]
				&{\includegraphics[width=1\linewidth]{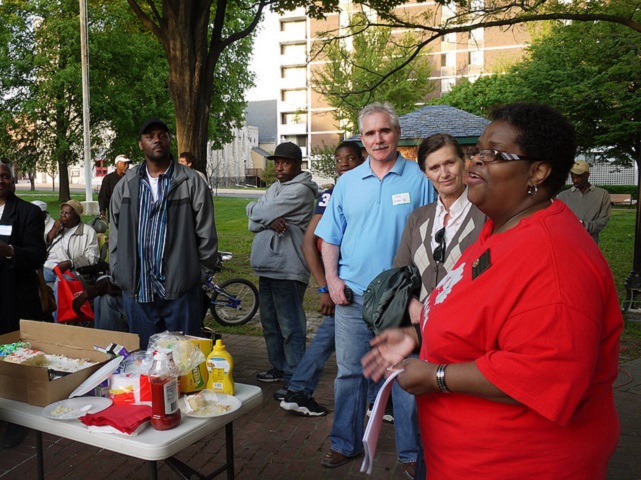}} & {\includegraphics[width=1\linewidth]{g.jpg}} \\
				\hline
				\multirow{2}*{Up-Down}&a group of people standing {\color{red} in a box of pizza}. & a man is {\color{red}jumping} in the air with a dog.\\
				\hline
				\multirow{2}*{CAAG-P} & a group of people standing around a table. & a man {\color{red}jumping} in the air with a frisbee. \\
				\hline
				\multirow{2}*{CAAG} &a group of people standing around a table with {\color{blue} food}. &a man {\color{red}jumping} in the air to catch a frisbee. \\
				
				\hline
		\end{tabular}}
		\caption{This table shows some examples generated by our methods respectively. The wrong phrases and fine-grained information are highlighted in red and blue separately.}
		\label{examples}
	\end{table}
	
	\subsection{Ablation Studies} 
	\subsubsection{Ablative Models} 
	We extensively explored different structures and settings of our method to gain insights about how and why it works. We selected Up-Down \cite{BTATD} captioner as the base model for ablation studies. Based on Up-Down, we devised the following variant models of CAAG: (1) \textbf{CAAG-P} only uses the enhanced primary network for inferencing. (2) \textbf{CAAG-C} replaces the objective function of CAAG with cross entropy, i.e. $A_t$ in Eq. (\ref{caag}) is a constant. (3)  \textbf{CAAG-H} replaces soft attention of CAAG with hard attention \cite{SAAT}. We also compare our models with Ruminant Decoding \cite{STP} that also considers a two-stage decoding mechanism.

	\subsubsection{Analysis} 
	
	Table \ref{ablation} shows the ablative results under our experimental settings. In general, all variant models of CAAG can significantly improve the base model, and can also outperform Base+RD model.  From the table, we also find the following observations: 
	\begin{itemize}
		\item Without using CAAG in the inference phase, the base model still be improved by 3.3 percent on CIDEr-D metric, which further verifies the effectiveness of our method. 
		\item CAAG and CAAG-H have better performance than CAAG-C because the advantage function can encourage correct generation and suppress bad samples. 
		\item CAAG is better than CAAG-H, which demonstrates that soft attention is able to capture more useful information from the global contexts compared with hard attention.
	\end{itemize}

	\begin{figure}[t]
		\begin{center}
			\includegraphics[width=0.7\linewidth]{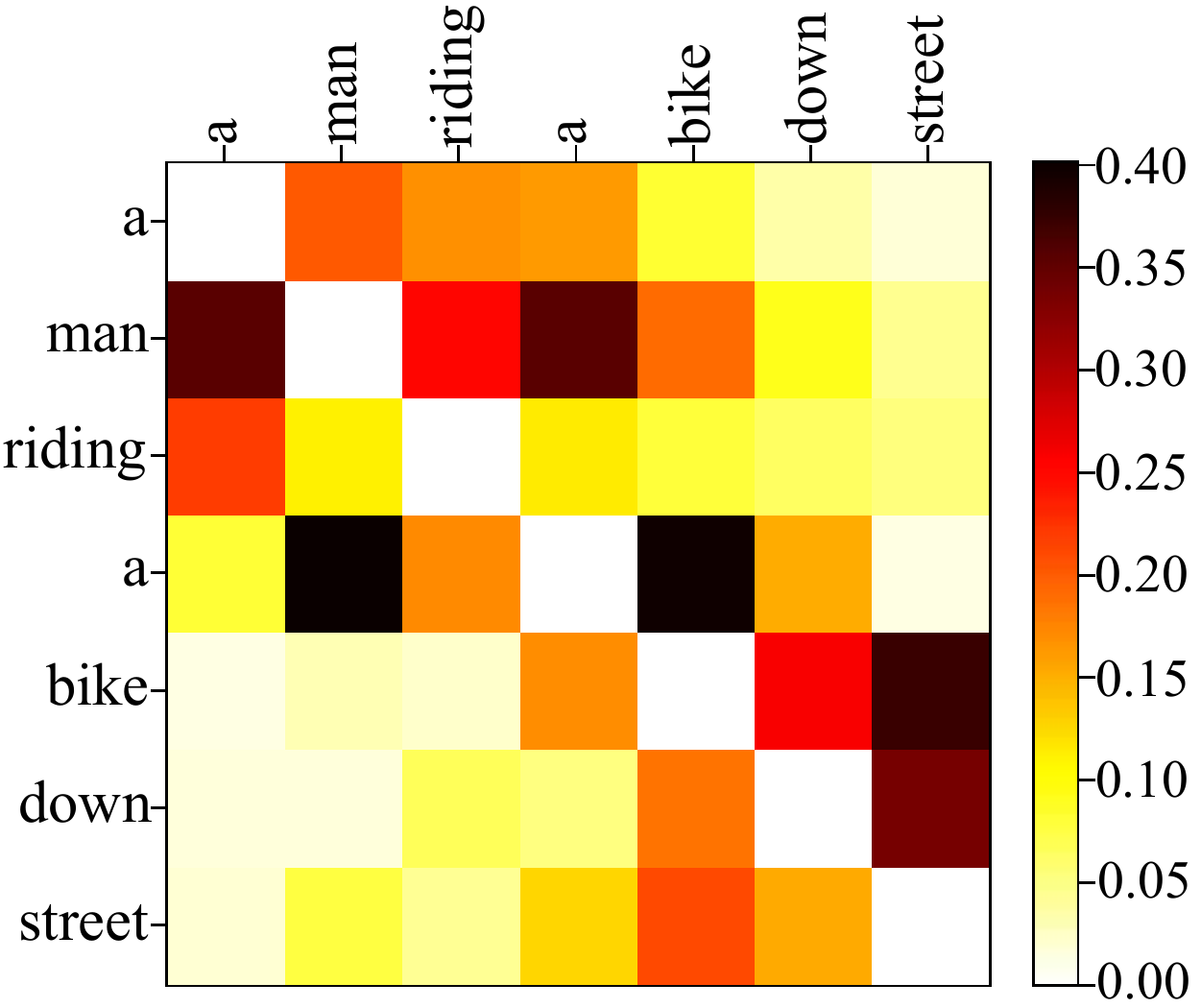}
		\end{center}
		\caption{An example of sematic attention weights visualization. The future generated token `bike' which has highest weight when predicting the target token `riding'.}
		\label{attention}
	\end{figure} 
	
	\begin{figure}[t]
		\centering
		\includegraphics[width=0.86\linewidth]{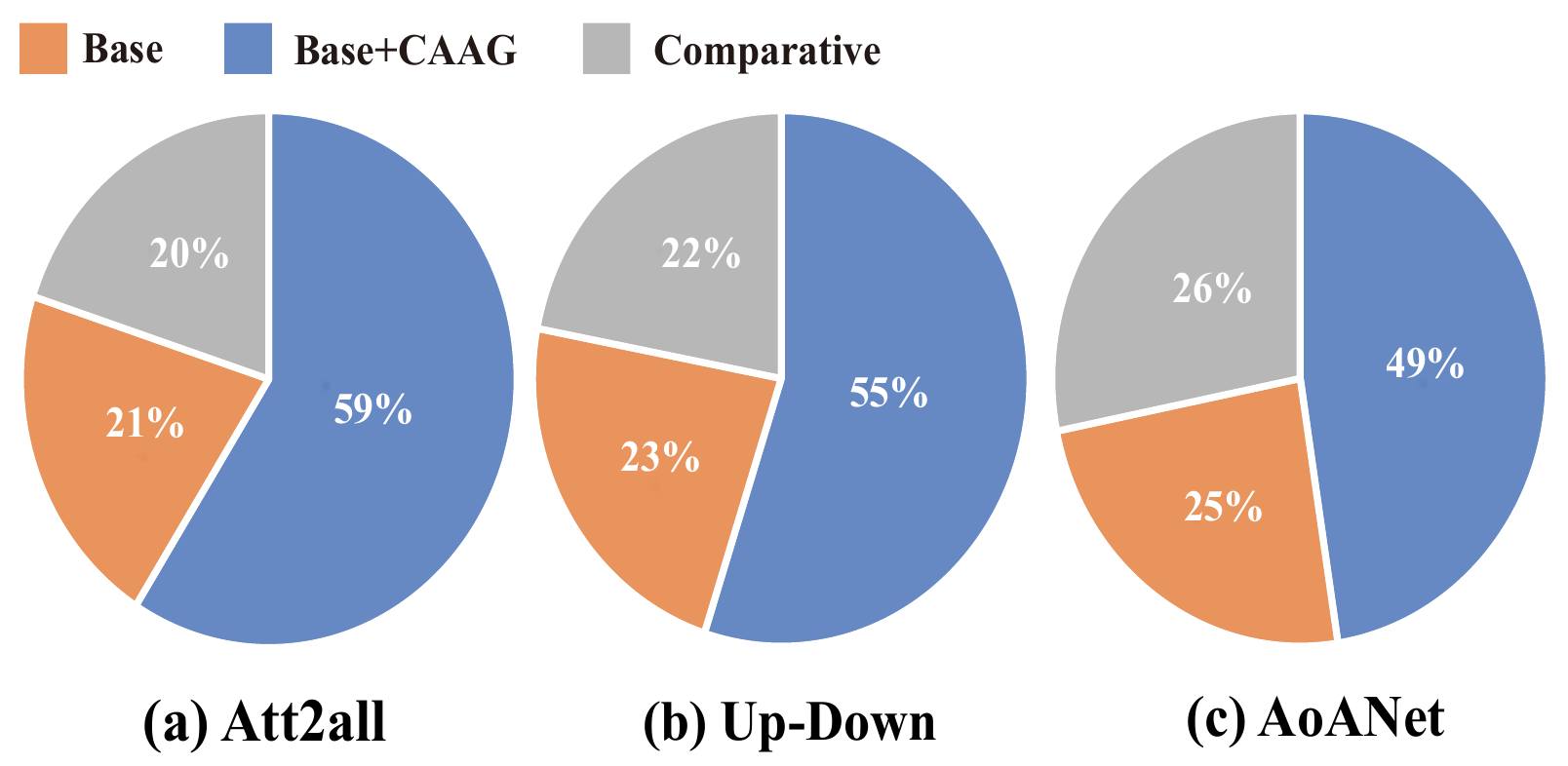}
		\caption{Results of human evaluation on 500 images randomly sampled from MSCOCO Karpathy test split. Each color indicates the percentage of workers who considers the capions generated by the corresponding model is more precise. And the gray color indicates that the two models are comparative.}
		\label{human}
	\end{figure}
	\begin{figure*}[t]
		\begin{center}
			\includegraphics[width=0.83\linewidth]{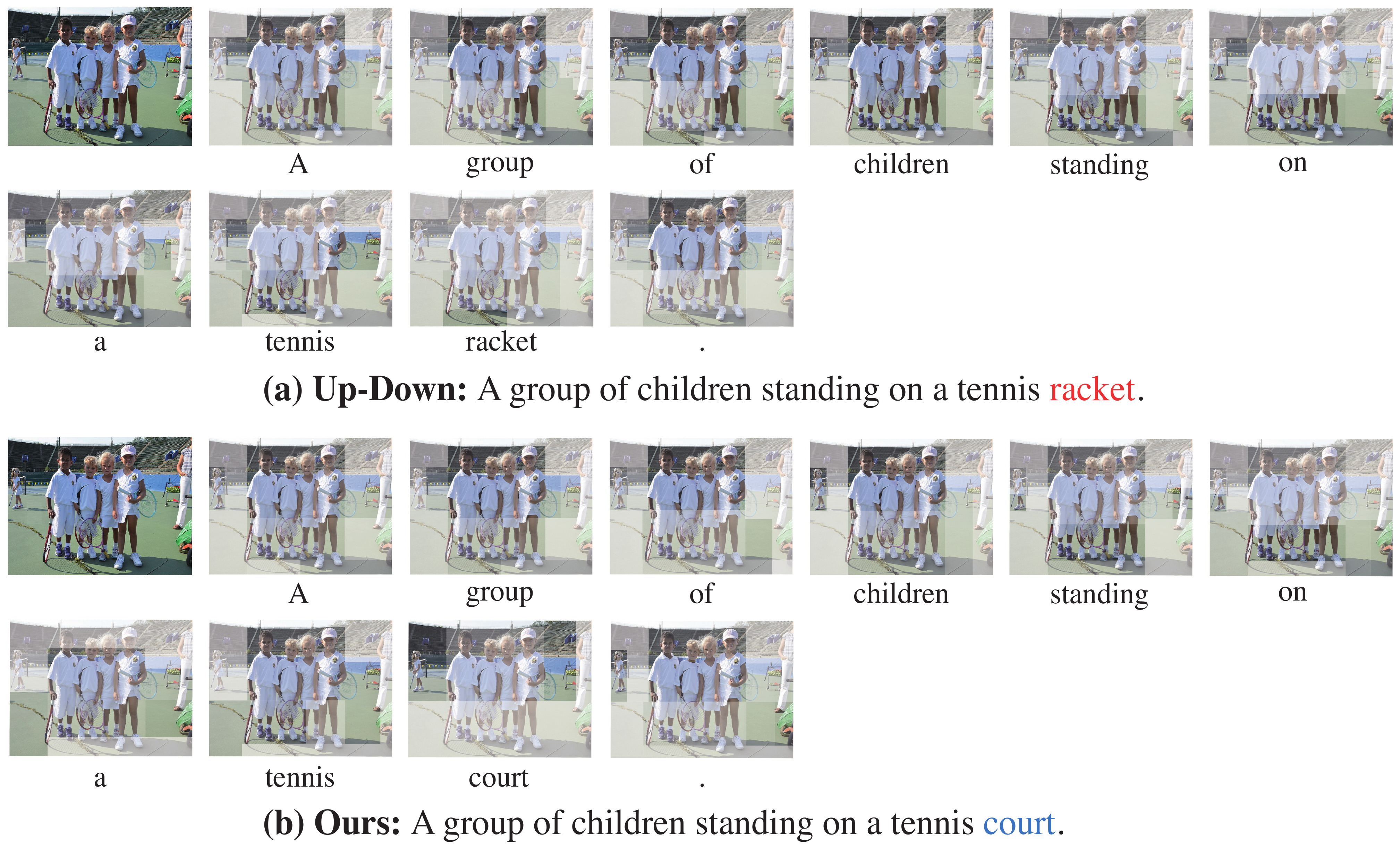}
		\end{center}
		\caption{Visualization of visual attention weights when generating image captions by Up-Down and our Up-Down+CAAG.}
		\label{att1}
	\end{figure*}
	\subsection{Qualitative Analysis}

	\subsubsection{Qualitative Examples} To validate the benefits of our proposed method, we conduct qualitative analysis over the image / caption pairs generated by Up-Down, Up-Down+CAAG-P and Up-Down+CAAG respectively. Table \ref{examples} provides some representative examples for comparisons.  In general, the captions generated by ours are better than base model. The two cases appearing above the table indicate that captions generated by Up-Down captioner are less descriptive or incorrect to some degree, e.g. ``with a table'' and ``refrigerator'', while our methods can generate captions with more precise information such as ``fork'' and ``microwave''. The left case appearing below suggests that the combination of primary network and CAAG can generate captions with more fine-grained information in the image, e.g. ``food'' and ``vegetables''. The last one shows a failure case, in which we predict a wrong action ``jumping''  for the ``man''. This indicates that more precise understanding of the image could alleviate such problem in the future.

	\subsubsection{Attention Visualization} Figure \ref{attention} provides an example to gain insights about how our CAAG takes advantage of global contexts.	As shown in the figure, we visualize the semantic attention weights $\beta_{i,t}$ from Eq. (\ref{location-weight}) to see which token in the global predictions (y-axis) is attended when reproducing the target word (x-axis) at every time step. Each column of the matrix in the plot indicates the weights associated  with  the  annotations. The white pixel means the weight of target token is zero. From the figure we can see that the future generated token ``bike'' which has highest probability is focused on when predicting the token ``riding''. In this way, CAAG can take advantage of future predictions to guide the captioning model to learn more complete semantics during training. Figure \ref{att1} gives an example of visualizing the visual attention weights for Up-Down and Up-Down+CAAG. From this figure we can find that our method with context-aware auxiliary guidance can attend to correct image regions when generating captions. In the example, the region  of ``racket" is attended by Up-Down model when generating the caption fragment ``A group of childern standing on a tennis", which is inconsistent with image content. On the contrary, our method can attend to correct regions and generate more precise caption.

	\subsection{Human Evaluation} 
	As the automatic evaluation metrics (e.g. BLEU and CIDEr-D) do not necessarily consistent with human judgment, we additionally conduct a user study to evaluate our methods against three baselines, i.e. Att2all, Up-Down and AoANet. We randomly sampled 500 images from Karpathy test split and invited 10 different workers who have prior experience with image captioning for human evaluation. We showed them two captions generated by base and base+CAAG models and asked them which one is more descriptive. The results of the comparisons on three different baselines are shown in Figure \ref{human}. From these pie charts, we can observe that when CAAG is used, the proportion of generated captions that are more descriptive is much higher than that of the baseline model.
	
	\section{Conclusion}
	In this paper, we propose Context-Aware Auxiliary Guidance (CAAG) mechanism to guide the captioning model (primary network) to learn complete semantics through reproducing the current generation based on the global contexts. As far as we know, CAAG is the first to take advantage of the global predictions in the process of caption generation. To validate the generality and adaptability of our proposed method, we employ CAAG to improve three popular attention based LSTM image captioners. Our model achieves competitive performance on the challenging Microsoft COCO image captioning benchmark.
	
	\section{ Acknowledgments}
	This work is supported by National Key R\&D Program No.2017YFB0803003, and the National Natural Science Foundation of China (No.61202226), We thank all anony- mous reviewers for their constructive comments.

	\bibliographystyle{aaai21}
	\bibliography{aaai21}

\begin{thebibliography}{44}
\providecommand{\natexlab}[1]{#1}
\providecommand{\url}[1]{\texttt{#1}}
\providecommand{\urlprefix}{URL }
\expandafter\ifx\csname urlstyle\endcsname\relax
  \providecommand{\doi}[1]{doi:\discretionary{}{}{}#1}\else
  \providecommand{\doi}{doi:\discretionary{}{}{}\begingroup
  \urlstyle{rm}\Url}\fi

\bibitem[{Anderson et~al.(2016)Anderson, Fernando, Johnson, and Gould}]{SPICE}
Anderson, P.; Fernando, B.; Johnson, M.; and Gould, S. 2016.
\newblock Spice: Semantic propositional image caption evaluation.
\newblock In \emph{European conference on computer vision}, 382--398. Springer.

\bibitem[{Anderson et~al.(2018)Anderson, He, Buehler, Teney, Johnson, Gould,
  and Zhang}]{BTATD}
Anderson, P.; He, X.; Buehler, C.; Teney, D.; Johnson, M.; Gould, S.; and
  Zhang, L. 2018.
\newblock Bottom-up and top-down attention for image captioning and visual
  question answering.
\newblock In \emph{Proceedings of the IEEE conference on computer vision and
  pattern recognition}, 6077--6086.

\bibitem[{Bahdanau, Cho, and Bengio(2015)}]{NMT}
Bahdanau, D.; Cho, K.; and Bengio, Y. 2015.
\newblock Neural Machine Translation by Jointly Learning to Align and
  Translate.
\newblock In \emph{International Conference on Learning Representations}.

\bibitem[{Banerjee and Lavie(2005)}]{METEOR}
Banerjee, S.; and Lavie, A. 2005.
\newblock METEOR: An automatic metric for MT evaluation with improved
  correlation with human judgments.
\newblock In \emph{Proceedings of the acl workshop on intrinsic and extrinsic
  evaluation measures for machine translation and/or summarization}, 65--72.

\bibitem[{Chen et~al.(2018)Chen, Ding, Zhao, and Han}]{TD}
Chen, H.; Ding, G.; Zhao, S.; and Han, J. 2018.
\newblock Temporal-difference learning with sampling baseline for image
  captioning.
\newblock In \emph{Thirty-Second AAAI Conference on Artificial Intelligence}.

\bibitem[{Chen et~al.(2017)Chen, Zhang, Xiao, Nie, Shao, Liu, and Chua}]{SCA}
Chen, L.; Zhang, H.; Xiao, J.; Nie, L.; Shao, J.; Liu, W.; and Chua, T.-S.
  2017.
\newblock Sca-cnn: Spatial and channel-wise attention in convolutional networks
  for image captioning.
\newblock In \emph{Proceedings of the IEEE conference on computer vision and
  pattern recognition}, 6298--6306. IEEE.

\bibitem[{Elliott and Keller(2013)}]{IDVD}
Elliott, D.; and Keller, F. 2013.
\newblock Image description using visual dependency representations.
\newblock In \emph{Proceedings of the 2013 Conference on Empirical Methods in
  Natural Language Processing}, 1292--1302.

\bibitem[{Erden and Tomiyama(2010)}]{HID}
Erden, M.~S.; and Tomiyama, T. 2010.
\newblock Human-Intent Detection and Physically Interactive Control of a Robot
  Without Force Sensors.
\newblock \emph{IEEE Transactions on Robotics} 26(2): 370--382.

\bibitem[{Fang et~al.(2015)Fang, Gupta, Iandola, Srivastava, Deng, Doll{\'a}r,
  Gao, He, Mitchell, Platt et~al.}]{FROM}
Fang, H.; Gupta, S.; Iandola, F.; Srivastava, R.~K.; Deng, L.; Doll{\'a}r, P.;
  Gao, J.; He, X.; Mitchell, M.; Platt, J.~C.; et~al. 2015.
\newblock From captions to visual concepts and back.
\newblock In \emph{Proceedings of the IEEE conference on computer vision and
  pattern recognition}, 1473--1482.

\bibitem[{Gao et~al.(2019)Gao, Wang, Wang, Ma, and Gao}]{SCSTN}
Gao, J.; Wang, S.; Wang, S.; Ma, S.; and Gao, W. 2019.
\newblock Self-critical n-step Training for Image Captioning.
\newblock \emph{Proceedings of the IEEE Conference on Computer Vision and
  Pattern Recognition} .

\bibitem[{Guo et~al.(2019)Guo, Liu, Lu, and Lu}]{STP}
Guo, L.; Liu, J.; Lu, S.; and Lu, H. 2019.
\newblock Show, tell and polish: Ruminant decoding for image captioning.
\newblock \emph{IEEE Transactions on Multimedia} .

\bibitem[{Herdade et~al.(2019)Herdade, Kappeler, Boakye, and
  Soares}]{transform}
Herdade, S.; Kappeler, A.; Boakye, K.; and Soares, J. 2019.
\newblock Image captioning: Transforming objects into words.
\newblock In \emph{Advances in Neural Information Processing Systems},
  11137--11147.

\bibitem[{Huang et~al.(2019)Huang, Wang, Chen, and Wei}]{AOA}
Huang, L.; Wang, W.; Chen, J.; and Wei, X.-Y. 2019.
\newblock Attention on attention for image captioning.
\newblock In \emph{Proceedings of the IEEE International Conference on Computer
  Vision}, 4634--4643.

\bibitem[{Karpathy, Joulin, and Li(2014)}]{ITS}
Karpathy, A.; Joulin, A.; and Li, F.~F. 2014.
\newblock Deep Fragment Embeddings for Bidirectional Image Sentence Mapping.
\newblock \emph{Advances in neural information processing systems} 3.

\bibitem[{Krishna et~al.(2017)Krishna, Zhu, Groth, Johnson, Hata, Kravitz,
  Chen, Kalantidis, Li, Shamma et~al.}]{VG}
Krishna, R.; Zhu, Y.; Groth, O.; Johnson, J.; Hata, K.; Kravitz, J.; Chen, S.;
  Kalantidis, Y.; Li, L.-J.; Shamma, D.~A.; et~al. 2017.
\newblock Visual genome: Connecting language and vision using crowdsourced
  dense image annotations.
\newblock \emph{International Journal of Computer Vision} 123(1): 32--73.

\bibitem[{Kuznetsova et~al.(2012)Kuznetsova, Ordonez, Berg, Berg, and
  Choi}]{CGN}
Kuznetsova, P.; Ordonez, V.; Berg, A.~C.; Berg, T.~L.; and Choi, Y. 2012.
\newblock Collective generation of natural image descriptions.
\newblock In \emph{Proceedings of the 50th Annual Meeting of the Association
  for Computational Linguistics: Long Papers-Volume 1}, 359--368. Association
  for Computational Linguistics.

\bibitem[{Li et~al.(2019)Li, Zhu, Liu, and Yang}]{ETAR}
Li, G.; Zhu, L.; Liu, P.; and Yang, Y. 2019.
\newblock Entangled transformer for image captioning.
\newblock In \emph{Proceedings of the IEEE International Conference on Computer
  Vision}, 8928--8937.

\bibitem[{Li et~al.(2011)Li, Kulkarni, Berg, Berg, and Choi}]{CSD}
Li, S.; Kulkarni, G.; Berg, T.~L.; Berg, A.~C.; and Choi, Y. 2011.
\newblock Composing simple image descriptions using web-scale n-grams.
\newblock In \emph{Proceedings of the Fifteenth Conference on Computational
  Natural Language Learning}, 220--228. Association for Computational
  Linguistics.

\bibitem[{Lin(2004)}]{ROUGE}
Lin, C.-Y. 2004.
\newblock Rouge: A package for automatic evaluation of summaries.
\newblock In \emph{Text summarization branches out}, 74--81.

\bibitem[{Lin et~al.(2014)Lin, Maire, Belongie, Hays, Perona, Ramanan,
  Doll{\'a}r, and Zitnick}]{MSCOCO}
Lin, T.-Y.; Maire, M.; Belongie, S.; Hays, J.; Perona, P.; Ramanan, D.;
  Doll{\'a}r, P.; and Zitnick, C.~L. 2014.
\newblock Microsoft coco: Common objects in context.
\newblock In \emph{European conference on computer vision}, 740--755. Springer.

\bibitem[{Liu et~al.(2018)Liu, Zha, Zhang, Zhang, and Wu}]{CAVP}
Liu, D.; Zha, Z.-J.; Zhang, H.; Zhang, Y.; and Wu, F. 2018.
\newblock Context-aware visual policy network for sequence-level image
  captioning.
\newblock \emph{Proceedings of the 26th ACM international conference on
  Multimedia} .

\bibitem[{Liu et~al.(2017)Liu, Zhu, Ye, Guadarrama, and Murphy}]{SPIDER}
Liu, S.; Zhu, Z.; Ye, N.; Guadarrama, S.; and Murphy, K. 2017.
\newblock Improved image captioning via policy gradient optimization of spider.
\newblock In \emph{Proceedings of the IEEE international conference on computer
  vision}, 873--881.

\bibitem[{Lu et~al.(2017)Lu, Xiong, Parikh, and Socher}]{knowing}
Lu, J.; Xiong, C.; Parikh, D.; and Socher, R. 2017.
\newblock Knowing when to look: Adaptive attention via a visual sentinel for
  image captioning.
\newblock In \emph{Proceedings of the IEEE conference on computer vision and
  pattern recognition}, 375--383.

\bibitem[{Lu et~al.(2018)Lu, Yang, Batra, and Parikh}]{NBT}
Lu, J.; Yang, J.; Batra, D.; and Parikh, D. 2018.
\newblock Neural Baby Talk.
\newblock In \emph{The IEEE Conference on Computer Vision and Pattern
  Recognition (CVPR)}.

\bibitem[{Mitchell et~al.(2012)Mitchell, Han, Dodge, Mensch, Goyal, Berg,
  Yamaguchi, Berg, Stratos, and Daum{\'e}~III}]{MID}
Mitchell, M.; Han, X.; Dodge, J.; Mensch, A.; Goyal, A.; Berg, A.; Yamaguchi,
  K.; Berg, T.; Stratos, K.; and Daum{\'e}~III, H. 2012.
\newblock Midge: Generating image descriptions from computer vision detections.
\newblock In \emph{Proceedings of the 13th Conference of the European Chapter
  of the Association for Computational Linguistics}, 747--756. Association for
  Computational Linguistics.

\bibitem[{Papineni et~al.(2002)Papineni, Roukos, Ward, and Zhu}]{BLEU}
Papineni, K.; Roukos, S.; Ward, T.; and Zhu, W.-J. 2002.
\newblock BLEU: a method for automatic evaluation of machine translation.
\newblock In \emph{Proceedings of the 40th annual meeting on association for
  computational linguistics}, 311--318.

\bibitem[{Qin et~al.(2019)Qin, Du, Zhang, and Lu}]{LBPF}
Qin, Y.; Du, J.; Zhang, Y.; and Lu, H. 2019.
\newblock Look Back and Predict Forward in Image Captioning.
\newblock In \emph{Proceedings of the IEEE Conference on Computer Vision and
  Pattern Recognition}, 8367--8375.

\bibitem[{Ranzato et~al.(2015)Ranzato, Chopra, Auli, and Zaremba}]{Exposure}
Ranzato, M.; Chopra, S.; Auli, M.; and Zaremba, W. 2015.
\newblock Sequence Level Training with Recurrent Neural Networks.
\newblock \emph{International Conference on Learning Representations} .

\bibitem[{Rennie et~al.(2017)Rennie, Marcheret, Mroueh, Ross, and Goel}]{SCST}
Rennie, S.~J.; Marcheret, E.; Mroueh, Y.; Ross, J.; and Goel, V. 2017.
\newblock Self-critical sequence training for image captioning.
\newblock In \emph{Proceedings of the IEEE Conference on Computer Vision and
  Pattern Recognition}, 7008--7024.

\bibitem[{Schmidt, Mael, and Wurtz(2006)}]{ASD}
Schmidt, P.; Mael, E.; and Wurtz, R.~P. 2006.
\newblock A sensor for dynamic tactile information with applications in
  human-robot interaction and object exploration.
\newblock \emph{Robotics and Autonomous Systems} 54(12): 1005--1014.

\bibitem[{Vedantam, Lawrence~Zitnick, and Parikh(2015)}]{CIDEr}
Vedantam, R.; Lawrence~Zitnick, C.; and Parikh, D. 2015.
\newblock Cider: Consensus-based image description evaluation.
\newblock In \emph{Proceedings of the IEEE conference on computer vision and
  pattern recognition}, 4566--4575.

\bibitem[{Vinyals et~al.(2015)Vinyals, Toshev, Bengio, and Erhan}]{SAT}
Vinyals, O.; Toshev, A.; Bengio, S.; and Erhan, D. 2015.
\newblock Show and tell: A neural image caption generator.
\newblock \emph{Proceedings of the IEEE Conference on Computer Vision and
  Pattern Recognition} 3156--3164.

\bibitem[{Vo et~al.(2019)Vo, Jiang, Sun, Murphy, Li, Feifei, and Hays}]{CTI}
Vo, N.; Jiang, L.; Sun, C.; Murphy, K.; Li, L.; Feifei, L.; and Hays, J. 2019.
\newblock Composing Text and Image for Image Retrieval - an Empirical Odyssey
  6439--6448.

\bibitem[{Wang et~al.(2019)Wang, Li, Huang, and Lazebnik}]{LTB}
Wang, L.; Li, Y.; Huang, J.; and Lazebnik, S. 2019.
\newblock Learning Two-Branch Neural Networks for Image-Text Matching Tasks.
\newblock \emph{IEEE Transactions on Pattern Analysis and Machine Intelligence}
  41(2): 394--407.

\bibitem[{Wang, Schwing, and Lazebnik(2017)}]{DA}
Wang, L.; Schwing, A.; and Lazebnik, S. 2017.
\newblock Diverse and Accurate Image Description Using a Variational
  Auto-Encoder with an Additive Gaussian Encoding Space.
\newblock In \emph{Advances in Neural Information Processing Systems 30},
  5756--5766.

\bibitem[{Williams(1992)}]{reinforce}
Williams, R.~J. 1992.
\newblock Simple statistical gradient-following algorithms for connectionist
  reinforcement learning.
\newblock \emph{Machine learning} 8(3-4): 229--256.

\bibitem[{Williams and Zipser(1989)}]{TF}
Williams, R.~J.; and Zipser, D. 1989.
\newblock A learning algorithm for continually running fully recurrent neural
  networks.
\newblock \emph{Neural computation} 1(2): 270--280.

\bibitem[{Wu et~al.(2018)Wu, Wang, Shen, Reid, and Hengel}]{NVD}
Wu, Q.; Wang, P.; Shen, C.; Reid, I.; and Hengel, A. 2018.
\newblock Are You Talking to Me? Reasoned Visual Dialog Generation Through
  Adversarial Learning.
\newblock In \emph{Proceedings of the IEEE Conference on Computer Vision and
  Pattern Recognition}, 6106--6115.

\bibitem[{Xu et~al.(2015)Xu, Ba, Kiros, Cho, Courville, Salakhudinov, Zemel,
  and Bengio}]{SAAT}
Xu, K.; Ba, J.; Kiros, R.; Cho, K.; Courville, A.; Salakhudinov, R.; Zemel, R.;
  and Bengio, Y. 2015.
\newblock Show, attend and tell: Neural image caption generation with visual
  attention.
\newblock In \emph{International Conference on Machine Learning}, 2048--2057.

\bibitem[{Yang et~al.(2019)Yang, Tang, Zhang, and Cai}]{AE}
Yang, X.; Tang, K.; Zhang, H.; and Cai, J. 2019.
\newblock Auto-encoding scene graphs for image captioning.
\newblock In \emph{Proceedings of the IEEE Conference on Computer Vision and
  Pattern Recognition}, 10685--10694.

\bibitem[{Yang et~al.(2016)Yang, Yuan, Wu, Cohen, and Salakhutdinov}]{RNCG}
Yang, Z.; Yuan, Y.; Wu, Y.; Cohen, W.~W.; and Salakhutdinov, R.~R. 2016.
\newblock Review Networks for Caption Generation.
\newblock In \emph{Advances in Neural Information Processing Systems 29},
  2361--2369.

\bibitem[{Yao et~al.(2018)Yao, Pan, Li, and Mei}]{explore}
Yao, T.; Pan, Y.; Li, Y.; and Mei, T. 2018.
\newblock Exploring visual relationship for image captioning.
\newblock In \emph{Proceedings of the European conference on computer vision
  (ECCV)}, 684--699.

\bibitem[{Yao et~al.(2019)Yao, Pan, Li, and Mei}]{HIP}
Yao, T.; Pan, Y.; Li, Y.; and Mei, T. 2019.
\newblock Hierarchy Parsing for Image Captioning.
\newblock In \emph{Proceedings of the IEEE/CVF International Conference on
  Computer Vision (ICCV)}.

\bibitem[{You et~al.(2016)You, Jin, Wang, Fang, and Luo}]{ICSA}
You, Q.; Jin, H.; Wang, Z.; Fang, C.; and Luo, J. 2016.
\newblock Image Captioning With Semantic Attention.
\newblock In \emph{The IEEE Conference on Computer Vision and Pattern
  Recognition (CVPR)}.

\end{thebibliography}
\end{document}